\documentclass[10pt,twocolumn,letterpaper]{article}

\usepackage{cvpr}              
\usepackage{stfloats}
\usepackage{pifont}

\usepackage{multirow}
\usepackage{makecell}
\usepackage{float}
\definecolor{mygreen}{HTML}{83C883}
\definecolor{myred}{HTML}{E28080}

\definecolor{cvprblue}{rgb}{0.21,0.49,0.74}
\usepackage[pagebackref,breaklinks,colorlinks,allcolors=cvprblue]{hyperref}

\def\confName{CVPR}
\def\confYear{2026}

\title{Fine-Grained Post-Training Quantization for Large Vision Language Models with Quantization-Aware Integrated Gradients}

\author{
 \textbf{Ziwei Xiang\textsuperscript{1,2}\thanks{Equal contribution. $^{\dag}$Corresponding to: ppyang, xyz @nlpr.ia.ac.cn}}~,
 \textbf{Fanhu Zeng\textsuperscript{1,2}\footnotemark[1]}~,
 \textbf{Hongjian Fang\textsuperscript{3}\footnotemark[1]}~,
 \textbf{Rui-Qi Wang\textsuperscript{4}},
 \textbf{Renxing Chen\textsuperscript{2}},\\
  \textbf{Yanan Zhu\textsuperscript{5}},
\textbf{Yi Chen\textsuperscript{1,2,6}},
\textbf{Peipei Yang\textsuperscript{1,2}$^{\dagger}$},
\textbf{Xu-Yao Zhang\textsuperscript{1,2}$^{\dagger}$}\\
 {\normalsize \textsuperscript{1}State Key Laboratory of Multimodal Artificial Intelligence Systems, CASIA}
 {\normalsize \textsuperscript{2}School of Artificial Intelligence, UCAS}  \\
  {\normalsize \textsuperscript{3} Beijing National Research Center for Information Science and Technology}  
{\normalsize \textsuperscript{4}Institute of Artificial Intelligence, USTB} \\
 {\normalsize \textsuperscript{5} School of Artificial Intelligence, Beihang University} 
 {\normalsize \textsuperscript{6}Zhongguancun Academy} \\
}
    
\begin{document}
\maketitle

\begin{abstract}
Large Vision Language Models~(LVLMs) have achieved remarkable success in a range of downstream tasks that require multimodal interaction, but their capabilities come with substantial computational and memory overhead, which hinders practical deployment. Among numerous acceleration techniques, post-training quantization is a popular and effective strategy for reducing memory cost and accelerating inference. However, existing LVLM quantization methods typically measure token sensitivity at the modality level, which fails to capture the complex cross-token interactions and falls short in quantitatively measuring the quantization error at the token level. As tokens interact within the model, the distinction between modalities gradually diminishes, suggesting the need for fine-grained calibration. Inspired by axiomatic attribution in mechanistic interpretability, we introduce a fine-grained quantization strategy on \textbf{Q}uantization-aware \textbf{I}ntegrated \textbf{G}radients~(\textbf{QIG}), which leverages integrated gradients to quantitatively evaluate token sensitivity and push the granularity from modality level to token level, reflecting both inter-modality and intra-modality dynamics. Extensive experiments on multiple LVLMs under both W4A8 and W3A16 settings show that our method improves accuracy across models and benchmarks with negligible latency overhead. For example, under 3-bit weight-only quantization, our method improves the average accuracy of LLaVA-onevision-7B by 1.60\%, reducing the gap to its full-precision counterpart to only 1.33\%. 
The code is available at \url{https://github.com/ucas-xiang/QIG}.
\end{abstract}

\section{Introduction}
Large Vision Language Models~(LVLMs)~\cite{bai2025qwen2,liu2024visual} have greatly advanced in recent years and exhibit astonishing performance across various downstream areas like image captioning~\cite{gurari2020captioning}, visual question answering~\cite{singh2019towards}, and so on~\cite{guo2025hide}. Meanwhile, the computation and latency scale steeply with model size, especially in the era where models with billions of parameters are commonplace, which limits their practical application in real-world scenarios. To address this, the main approaches include pruning~\cite{kong2025token, zeng2024m2m, rao2021dynamicvit}, distillation~\cite{hinton2015distilling}, and quantization~\cite{frantar2022gptq, lin2024awq}. Among them, post-training quantization~(PTQ)~\cite{lin2021fq, frantar2022gptq, zhao2024vidit} provides a feasible approach to accelerate inference. By applying weight-only or weight-activation quantization, it reduces memory usage and computation overload while minimizing reconstruction error with a small calibration set, thereby maintaining task performance and achieving strong accuracy–efficiency trade-offs in a training-free manner.

\begin{figure}
    \centering
    \includegraphics[width=1.0\linewidth]{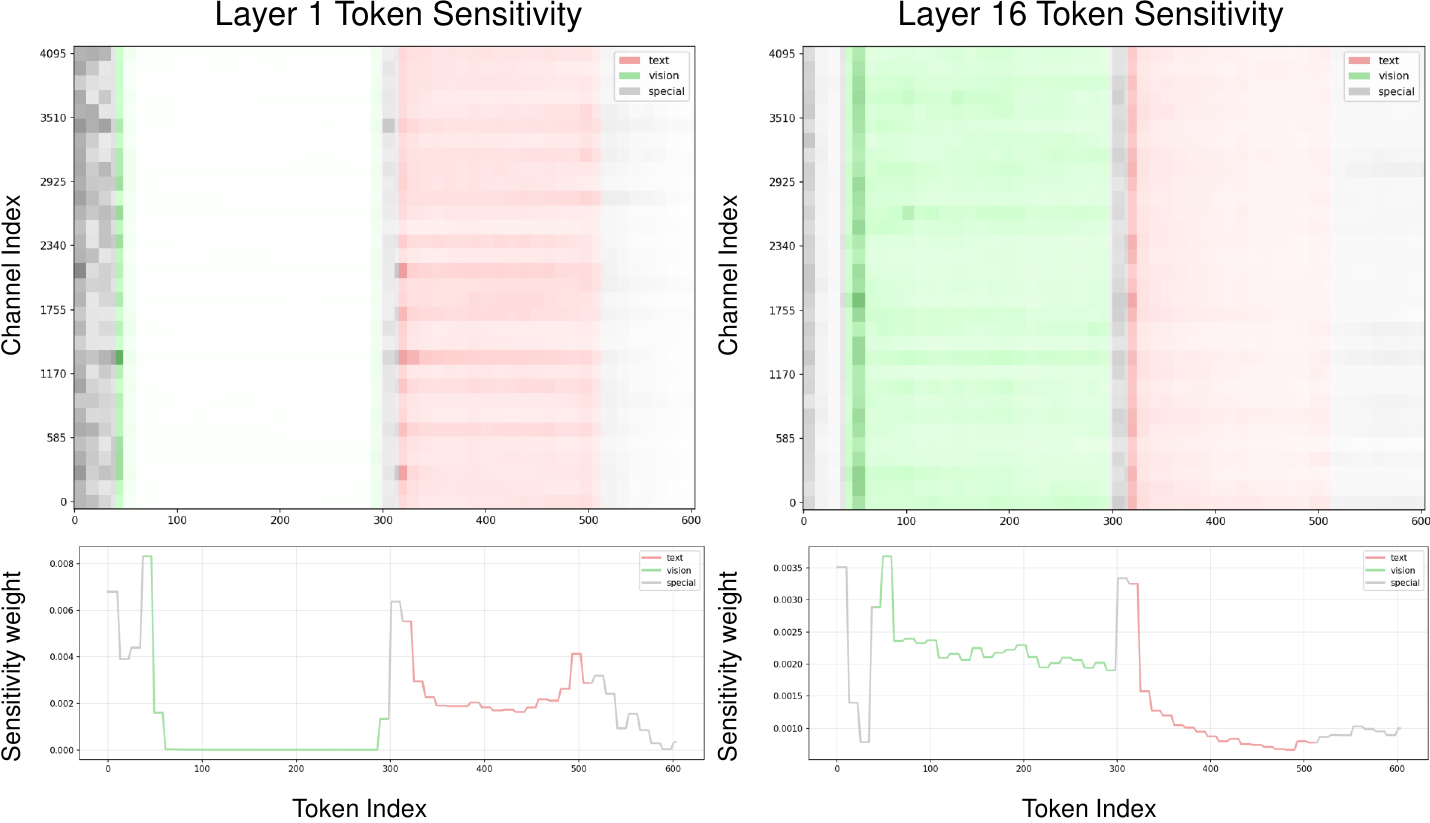}
    \caption{
    \textbf{Token-level quantization sensitivity across layers in the form of heatmap and curves.}
    At layers 1 and 16, we show both the token-level sensitivity
    \emph{heatmap} and its \emph{channel-averaged} line curve for \textcolor{gray}{special},
    \textcolor{mygreen}{vision}, and \textcolor{myred}{text} tokens, measured using our
    Quantization-aware Integrated Gradients~(QIG). 
    }
    \label{fig:sensitivity}
    \vspace{-20pt}
\end{figure}

Quantization has made great progress in large language models for efficient inference~\cite{zeng2025token, wu2023ppt}, with techniques such as rotation~\cite{lin2024duquant} and channel scaling~\cite{lin2024awq}. Building on these advances, recent LVLM quantization methods exploit multimodal structure to improve performance~\cite{li2025mbq, xie2024advancing, zeng2025modalprompt}. MBQ~\cite{li2025mbq} introduces a gradient-based objective that reweights reconstruction errors across modalities, mitigating inter-modality imbalance. QSLAW~\cite{xie2024advancing} designs a quantization-aware scale learning framework with a multimodal warmup for efficient instruction tuning. Q-VLM~\cite{wang2024qvlm} performs block-level joint optimization guided by activation entropy to reduce greedy mismatch.

Despite the great progress in LVLM quantization, several issues remain to be tackled. (1) The complex interaction between modalities makes the distribution largely vary in different layers and modalities. As illustrated in  Fig.~\ref{fig:sensitivity}, token sensitivity differs not only between modalities~(inter-modality) but also within a modality~(intra-modality) and across depth, suggesting that modality-level quantization is insufficient to capture token-wise dynamics in LVLMs; (2) There remains a gap between the quantized model and the original model. This naturally calls for a fine-grained analysis of how each token contributes to quantization-induced output perturbations. Existing methods avoid token-level analysis, which may be attributed to the weak correlation between common proxies such as attention and the true quantization error, as well as their tendency to overlook the most influential tokens. This limitation underscores the need for a direct and effective way to define token-level sensitivity for PTQ.

Motivated by this, we aim to explore fine-grained LVLM quantization and push the quantitative measurement of granularity from the modality level to the token level. We draw on the concept of axiomatic attribution~\cite{ancona2018towards} from mechanistic interpretability~\cite{bereska2024mechanistic}, which enables us to effectively analyze the perturbation sensitivity of each token by calculating the integrated gradients~\cite{sundararajan2017axiomatic} during calibration. Concretely, we calculate the Quantization-aware Integrated Gradients~(QIG) from the quantized reference input to the actual input, thereby obtaining a token-level sensitivity score that quantifies the influence of each input token on the final model quantization error\footnote{The completeness property of quantization-aware integrated gradients is proved in Appendix~\ref{sec:qig-completeness}.}. Additionally,  we further apply a robust IQR-based clipping to suppress extreme token importance values and stabilize the sensitivity estimation during quantization. Empirically, QIG strongly correlated with actual quantization errors, validating its suitability as a proxy signal for guiding fine-grained quantization.

We conduct comprehensive experiments on multiple open-source LVLMs for both weight-only and weight-activation quantization. The results show that our method delivers consistent gains on various multimodal benchmarks. For example, under 3-bit weight-only quantization, our method improves the average accuracy of LLaVA-onevision-7B by 1.60\%, reducing the gap to its full-precision counterpart to only 1.33\%. These results demonstrate that our method can significantly improve the accuracy of quantized LVLMs with negligible latency overhead, highlighting its practical efficiency. Our main contributions are summarized as follows:

\begin{itemize}
  \item We reveal the complex interaction between modalities in LVLM quantization, highlighting the necessity of fine-grained sensitivity measurements for multimodal inputs.
  \item We introduce the concept of axiomatic attribution and develop Quantization-aware Integrated Gradients, a quantization-specific sensitivity estimation method that provides token-level attributions of quantization error and directly guides fine-grained post-training quantization.

  \item We conduct extensive experiments on various multimodal benchmarks to comprehensively demonstrate the superiority and effectiveness of our method.
\end{itemize}

\begin{figure*}[t]
    \centering
    \includegraphics[width=1.0\textwidth]{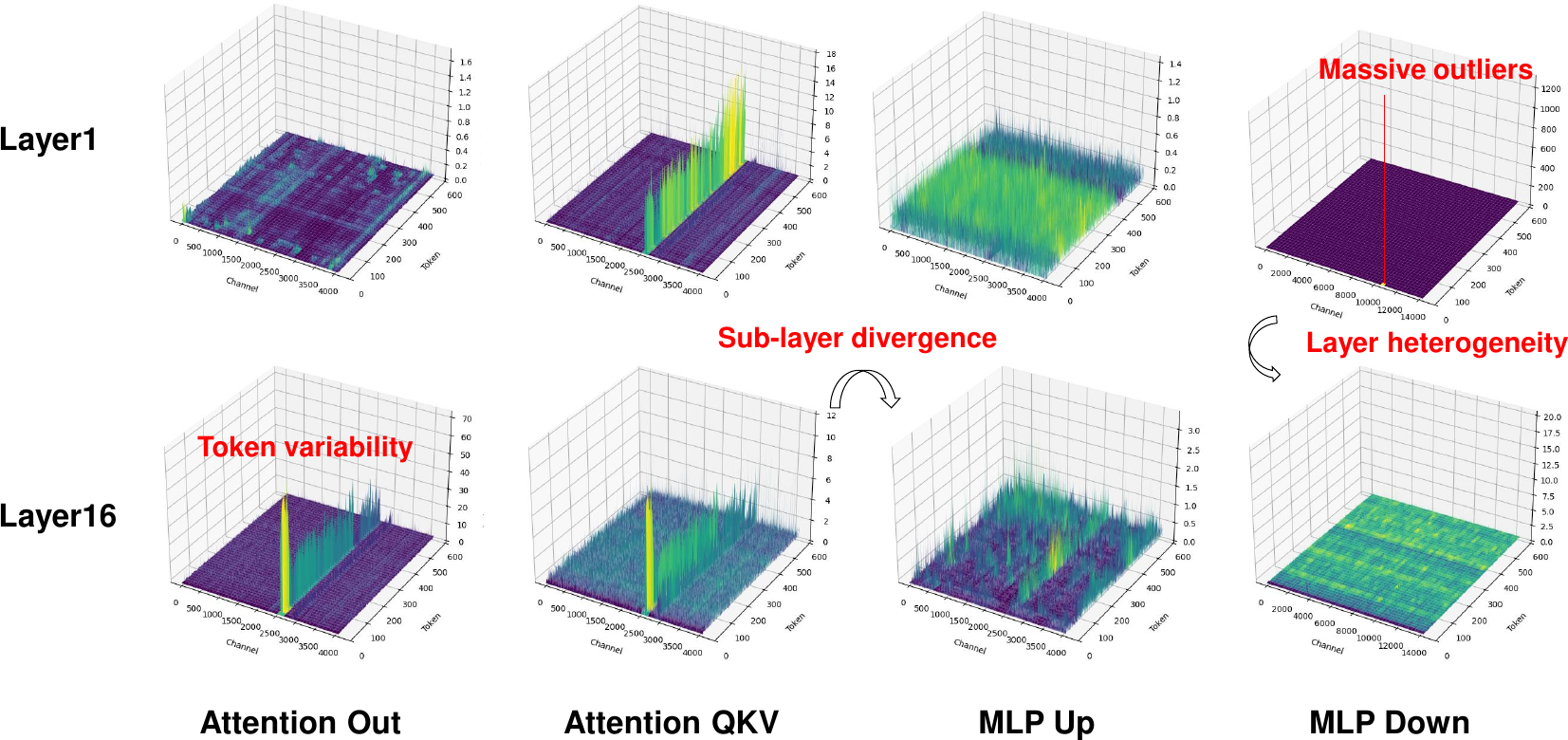}  
    \caption{Visualization of activation distributions in InternVL2-8B during calibration. We visualize two representative layers and four linear sub-layers. In each panel, the horizontal axis denotes token positions in the multimodal sequence and the vertical axis indexes hidden channels; color encodes the average activation magnitude per token–channel pair over the calibration set. The plots reveal four recurring phenomena: massive activations, layer heterogeneity, sub-layer divergence, and token variability. These patterns indicate that coarse modality-level sensitivity modeling is insufficient, motivating our token-level sensitivity weighting.}    
      \label{fig:activation}
    \vspace{-8pt}
\end{figure*}

\section{Related Work}
\subsection{Large Vision Language Models}
Large vision language models  bridge vision and language by projecting image features into the  Large Language Models~(LLMs) input space~\cite{li2023blip,awadalla2023openflamingo, zeng2025robustmerge}. Representative architectures such as LLaVA~\cite{li2024llava}, InternVL~\cite{chen2024internvl}, and Qwen-VL~\cite{bai2025qwen2} encode an image using a Vision Transformer ~\cite{dosovitskiy2020image} or CLIP encoder~\cite{radford2021learning} into a sequence of visual patch tokens. These visual tokens are then combined with text tokens and task-specific special tokens (\eg, \texttt{<bos>}, \texttt{<eos>}, and the \texttt{<image>} token, which demarcates visual content) into a unified input sequence. This heterogeneous multimodal sequence allows the LLM to process and reason over information from both modalities ~\cite{chen2024internvl, bai2025qwen2, li2024llava}. Unlike works proposing new architectures for better modality alignment, we focus on efficient acceleration for LVLMs.

\subsection{Post-Training Quantization}
Post-training quantization~(PTQ)~\cite{gholami2022survey,shao2025tr,zheng2025first} is a widely adopted compression technique that converts full-precision weights and activations into lower-bit representations without requiring retraining. In LLMS, several representative PTQ approaches have been proposed~\cite{lin2024awq,xiao2023smoothquant,lin2024duquant}. RTN applies simple rounding-to-nearest quantization as a strong baseline, AWQ~\cite{lin2024awq} introduces activation-aware weight quantization to preserve salient channels, GPTQ~\cite{frantar2022gptq} minimizes layer-wise reconstruction error through second-order approximation, and SmoothQuant~\cite{xiao2023smoothquant} balances activation and weight ranges to stabilize quantization during inference. Recently, PTQ has been extended to LVLMs to reduce their multimodal inference cost~\cite{li2025mbq,yu2025activation}. However, existing works mainly aim to achieve balanced quantization across modalities or layers, while the uneven token-wise sensitivity within each layer remains largely underexplored.

\subsection{Interpretability and Token Sensitivity}
Interpretability research aims to elucidate how the internal components of deep models interact to produce specific behaviors~\cite{lin2025survey}, offering a causal understanding beyond input–output correlations~\cite{ben2024lvlm}.  Intervention-based methods analyze model behavior by modifying inputs or intermediate activations. Occlusion Sensitivity~\cite{zeiler2014visualizing} measures the influence of each input region on model predictions by systematically occluding local areas of the input, while Activation Patching~\cite{zhang2023towards} examines causal mediation within models by substituting activations between corrupted and clean forward passes. In contrast, gradient-based methods estimate feature importance using gradient information, such as Integrated Gradients~(IG)~\cite{sundararajan2017axiomatic} and SmoothGrad~\cite{smilkov2017smoothgrad}. Although these approaches have achieved notable success in model analysis and visualization, most studies remain centered on interpretability itself rather than directly exploiting interpretability signals for model optimization.

\section{Method}
\subsection{Preliminaries}
\label{sec:prelim}
Existing PTQ methods automatically search for optimal quantization hyperparameters by minimizing the reconstruction error of each transformer block during a calibration process. Building on reconstruction-aware calibration, recent weight–activation~(WA) PTQ approaches~\cite{xiao2023smoothquant,li2025mbq} aim to quantize both weights and activations to low precision while maintaining model quality. To alleviate the large quantization error caused by activation outliers, these methods perform channel-wise equalization~(CWE)~\cite{xiao2023smoothquant} on both the weight and activation matrices.

Let $\mathbf{X} = [\mathbf{X}_1, \ldots, \mathbf{X}_T] \in \mathbb{R}^{d \times T}$ denote the activation matrix of a transformer block, where each column $\mathbf{X}_i \in \mathbb{R}^{d}$ is the embedding of the $i$-th token in a sequence of length $T$. Let $\mathbf{W} \in \mathbb{R}^{m \times d}$ be the weight matrix of a linear sub-layer, where $m$ denotes the output dimension of this linear sub-layer. Let $\mathbf{E} \in \mathbb{R}^{d}$ denote the channel-wise scaling factors applied along the hidden dimension $d$. We use ``$*$'' to denote channel-wise~(per-channel) scaling of $\mathbf{W}$ and $\mathbf{X}$ by $\mathbf{E}$. Specifically, CWE searches for optimal scaling factors $\mathbf{E}$ by minimizing the mean squared error~(MSE) between the quantized and original outputs of each transformer block. The optimization objective for weight-activation quantization can be formulated as:
{\small
\begin{equation}
\label{eq:cwe_wa}
    \mathbf{E}^* =
    \mathop{\mathrm{arg\,min}}\limits_{\mathbf{E}}
    \left\|
        Q_W(\mathbf{W} * \mathbf{E}) \,
        Q_X(\mathbf{E}^{-1} * \mathbf{X})
        - \mathbf{W}\mathbf{X}
    \right\|_2^2,
\end{equation}
}
where $Q_W(\cdot)$ and $Q_X(\cdot)$ denote the quantization functions for weights and activations, respectively.

This formulation aims to jointly optimize the scaling of weights and activations, ensuring that quantization preserves the representational capacity of each transformer block. For simplicity, we use \textbf{WxAy} to indicate the quantization format, where \textbf{x} and \textbf{y} represent the bit-widths for weight and activation, respectively. For example, W4A8 denotes quantizing weights to 4 bits and activations to 8 bits.

\subsection{Sensitivity Differences Between Modalities and Tokens}
Quantization sensitivity characterizes the degree to which a token or layer is affected by quantization noise. Since the dynamic range of activations determines the quantization scaling factor, activation statistics provide a practical proxy for estimating sensitivity. Therefore, before estimating sensitivity explicitly, we first analyze activation distributions to understand the origins of sensitivity differences. 

From Fig.~\ref{fig:activation}, we observe four recurring phenomena across two layers~(Layer1 and Layer16) and four linear sub-layers~(Attention Out, Attention QKV, MLP Up, MLP Down): (i) \textbf{Massive outliers}, large activation outliers persist across layers, forcing quantizers to widen the dynamic range; (ii) \textbf{Layer heterogeneity}, different Transformer layers display distinct activation behaviors; (iii) \textbf{Sub-layer divergence}, even within the same Transformer block, different sub-layers exhibit heterogeneous activation characteristics; and (iv) \textbf{Token variability}, within the same sub-layer, activations vary substantially across tokens, causing quantization to affect different tokens unevenly. These findings reveal that quantization sensitivity is not only modality-dependent~(vision vs.\ language) but also highly token-dependent. However, existing LVLM quantization methods model sensitivity only at the modality level and implicitly assume equal 
sensitivity for all tokens within a modality. We hypothesize that overlooking token-level sensitivity variations fundamentally limits the performance of current LVLM quantization strategies.

\begin{table}[t]
\centering
\resizebox{0.95\linewidth}{!}{
\begin{tabular}{lcc}
\toprule
\textbf{Sensitivity Type} & \textbf{Granularity} & \textbf{Accuracy~(\%)} \\
\midrule
\multirow{3}{*}{\textbf{Gradient-based}} 
    & Modality-level       & 57.36 \\
    & Token-level                & 55.78 \\
    & Token-level~(+ special)    & 55.65 \\
\midrule
\multirow{3}{*}{\textbf{Attention-based}}
    & Modality-level             & 56.43 \\
    & Token-level                & 57.12 \\
    & Token-level~(+ special)    & 57.52 \\
\midrule
\multirow{2}{*}{\textbf{Perturbation-based}}
    & Modality-level             & 56.81 \\
    & Token-level~(+ special)    & \textbf{57.72} \\
\bottomrule
\end{tabular}}
\caption{
Comparison of \textbf{modality-level} and \textbf{token-level} sensitivity estimation strategies on VizWiz~(W4A8, InternVL2-8B).
}
\label{tab:token_sensitivity}
\vspace{-20pt}
\end{table}

To examine whether fine-grained sensitivity modeling is necessary, we run controlled experiments on InternVL2-8B~(W4A8), keeping all quantization hyperparameters and calibration data fixed and varying only the sensitivity estimation strategy. We compare three approaches:

\begin{itemize}
    \item \textbf{Gradient-based sensitivity.}
    Following MBQ~\cite{li2025mbq}, sensitivity is estimated from gradients of the supervised fine-tuning~(SFT) loss. 
    At the modality level, one sensitivity value is assigned to visual tokens and one to textual tokens. 
    At the token level, each token~(vision, text, and special) receives an individual score.

    \item \textbf{Attention-based sensitivity.}
    Sensitivity is derived from attention scores. 
    Modality-level sensitivity aggregates scores within each modality, 
    while token-level sensitivity directly uses per-token attention statistics.

    \item \textbf{Perturbation-based sensitivity.}
    Sensitivity is obtained by perturbing tokens and measuring the change of block's outputs. 
    Modality-level sensitivity jointly perturbs all visual or all textual tokens, 
    whereas token-level sensitivity uses a leave-one-out scheme over individual tokens.
\end{itemize}

Tab.~\ref{tab:token_sensitivity} shows three trends. (1) \textbf{Gradient-based token-level weighting performs worse than modality-level,}
showing that SFT gradients do not correlate with quantization sensitivity. Once quantization noise is introduced, the gradient distribution changes, and the mismatch accumulates over depth. (2) \textbf{Attention-based sensitivity gives only modest and unstable gains}, which is consistent with the attention-sink phenomenon~\cite{kang2025see}, where certain tokens receive spuriously high attention. (3) \textbf{Perturbation-based sensitivity performs best}, as it directly measures the model's response to quantization noise, but it requires repeated forward passes and is computationally expensive.

These observations suggest that token-level sensitivity can improve quantization when it is estimated accurately, yet gradient- and attention-based proxies are misaligned with quantization error, and perturbation-based estimation is too costly to use directly. 
This motivates the fine-grained quantization method introduced in the next section.

\subsection{Fine-Grained Quantization}
\label{sec:main_method}

\begin{figure}[t]
    \centering
    \includegraphics[width=1.0\linewidth]{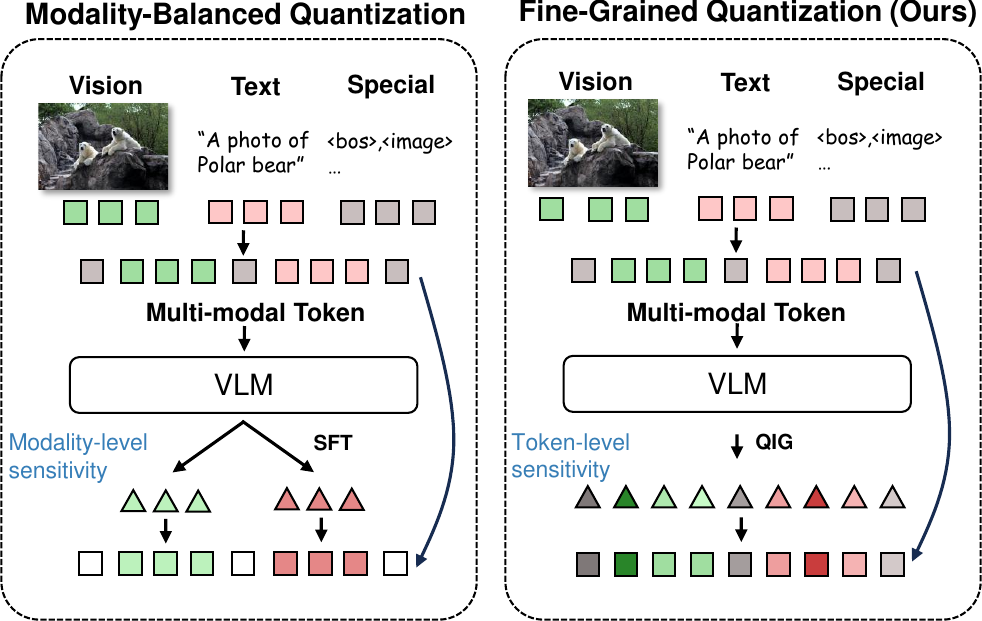}
    \caption{Comparison between modality-balanced quantization and our fine-grained quantization. Different colors indicate token types. Unlike MBQ, which assigns modality-level sensitivity, our method computes token-level sensitivity via Quantization-aware Integrated Gradients~(QIG) during calibration, enabling more effective quantization.}
    \label{fig:method}
    \vspace{-10pt}
\end{figure}

Building on this analysis, we propose our fine-grained method. As illustrated in Fig.~\ref{fig:method}, prior modality-based PTQ methods assign uniform sensitivity weights to all tokens within a modality. However, token-level sensitivity is highly heterogeneous, varying across tokens, layers, and architectures. Modality-level weighting fails to capture this granularity, leading to suboptimal quantization. To address this, we introduce a token-level sensitivity estimator that adaptively prioritizes more vulnerable tokens during calibration, improving overall quantization quality. We term this fine-grained quantization.

Motivated by interpretability and attribution principles, we draw on axiomatic attribution~\cite{ancona2018towards}, which naturally quantifies each token’s contribution to model behavior and thus serves as a suitable foundation for measuring token importance during quantization. We start from the classical Integrated Gradients~(IG)~\cite{sundararajan2017axiomatic}, which measures the cumulative contribution of each token along the straight path from a reference input $x'$ to the actual input $x$, where $f(\cdot,\cdot)$ denotes the output of the block:

\begin{equation}
\label{eq:ig}
IG(x) = (x - x^{'}) \int_{0}^{1} \frac{\partial f(x_{\alpha}, w)}{\partial x_{\alpha}} \, d\alpha,
\end{equation}
where $x_{\alpha} = x^{'} + \alpha (x - x^{'})$ and $f(\cdot, w)$ denotes the full-precision model. Eq.~(\ref{eq:ig}) reflects token contributions to the full-precision prediction; however, it does not reveal how sensitive the quantization-induced error is to each token.

To align the attribution with quantization, we instead explain the output gap between the full-precision model and the quantized model. Let $x^{q}$ denote the reference input along the attribution path and let $w^{q}$ be the quantized weights. In our main setting of joint weight–activation quantization, $x^{q}$ corresponds to the quantized input; in the case of weight-only quantization, activations remain in full precision and $x^{q}$ reduces to the zero baseline. At this step, we shift the IG objective from attributing the model’s absolute prediction to attributing the prediction difference caused by quantization, allowing us to isolate the impact of quantization errors. We define the token-level Quantization-aware Integrated Gradients~(QIG) as:
{\small
\begin{equation}
\label{eq:qig}
QIG(x) = (x - x^{q}) \int_{0}^{1} 
\frac{\partial \left( f(x_{\alpha}, w) - f(x_{\alpha}, w^{q}) \right)}{\partial x_{\alpha}} 
\, d\alpha,
\end{equation}
}
with $x_{\alpha} = x^{q} + \alpha (x - x^{q})$. 
Here, $QIG(x)$ is a token-wise attribution vector, and $QIG_i(x)$ denotes the attribution score of the $i$-th token, quantifying how much restoring that token from its quantized representation reduces the output discrepancy between $f(x, w)$ and $f(x, w^{q})$. Intuitively, a token with a large $QIG$ has a disproportionately strong influence on the quantization error. Small perturbations in this token’s embedding can significantly alter the output discrepancy between $f(x, w)$ and $f(x, w^{q})$. Compared to $IG$, $QIG$ is directly tied to the error that actually appears in PTQ, and it also satisfies a completeness property analogous to IG, for which we provide a formal derivation in Appendix~\ref{sec:qig-completeness}.

However, raw $QIG$ values are often heavy-tailed, causing a few extreme tokens to dominate optimization. To suppress such outliers while preserving relative importance, we apply interquartile range~(IQR) clipping~\cite{chatfield1986exploratory} to obtain the clipped score:
\begin{equation}
\scalebox{0.82}{$
C(QIG_i) = \operatorname{clip}\left(
    QIG_i,\;
    Q_1 - 1.5 \cdot IQR,\;
    Q_3 + 1.5 \cdot IQR
\right)
$}
\end{equation}
where $Q_1$ and $Q_3$ are the first and third quartiles, and $IQR = Q_3 - Q_1$. We then normalize these scores to obtain the token importance coefficients:
\begin{equation}
\label{eq:tok_weight}
\lambda_i = \frac{C(QIG_i)}{\sum_{j=1}^{T} C(QIG_j)},
\end{equation}
ensuring that the coefficients sum to one.

We integrate QIG into CWE to optimize the equalization factors. Keeping the WA quantization scheme in Eq.~(\ref{eq:cwe_wa}) unchanged, we reweight each token’s reconstruction error by its importance score $\lambda_i$. The objective function becomes:
\begin{equation}
\label{eq:cwe_wa_tok}
\scalebox{0.82}{$
\mathbf{E}^* = \arg\min_{\mathbf{E}} \sum_{i=1}^{T} 
\lambda_i \, \big\|
 Q_W(\mathbf{W} * \mathbf{E}) \, Q_X(\mathbf{E}^{-1} * \mathbf{X}_i) 
- \mathbf{W}\mathbf{X}_i
\big\|_2^2.
$}
\end{equation}
where $X_i$ represent the $i$-th input token activation of each linear layer. For weight-only quantization, it becomes:
\begin{equation}
\label{eq:cwe_w_tok}
\scalebox{0.82}{$
\mathbf{E}^* = \arg\min_{\mathbf{E}} \sum_{i=1}^{T} 
\lambda_i \, \big\|
 Q_W(\mathbf{W} * \mathbf{E}) \, (\mathbf{E}^{-1} * \mathbf{X}_i) 
- \mathbf{W}\mathbf{X}_i
\big\|_2^2.
$}
\end{equation}

In this way, the scale search is biased towards tokens that are empirically more sensitive to quantization, while the overall CWE framework remains unchanged. Beyond offering a more fine-grained, token-level sensitivity analysis, our approach improves performance while introducing virtually no additional computational cost.

\label{sec:method}
\section{Experiment}
\subsection{Experimental Setup}
\label{sec:Setup}
\noindent \textbf{Implementation Details.}
In line with prior studies~\cite{li2025mbq,xiao2023smoothquant,lin2024awq}, we apply per-token activation quantization and per-channel weight quantization. Given that W8A8 quantization has been established as lossless in precision by SmoothQuant~\cite{xiao2023smoothquant}, our primary evaluation in this paper focuses on W4A8 and W3A16. All experiments are conducted on a single NVIDIA A800 GPU~(80GB).

\noindent \textbf{Calibration Datasets.} 
Following prior work, we adopt the improved COCO Caption dataset from ShareGPT4V~\cite{sharegpt4v} and randomly sample 128 image–caption pairs for calibration. Each pair is formatted according to the conversational prompt style of the target LVLM.

\noindent \textbf{Models.}
We conduct both W3A16 and W4A8 quantization on numerous leading open-source LVLMs, including LLaVA-onevision-7B~\cite{li2024llava}, Qwen2-VL-7B~\cite{Qwen2VL}, and InternVL2-8B/26B~\cite{chen2024internvl}. For the LLaVA-onevision series, we select versions that adopt Qwen2 as the language model backbone and SigLIP-400M~\cite{zhai2023sigmoid} as the vision encoder.

\noindent \textbf{Baselines.}
For weight-only quantization, we compare our method with vanilla round-to-nearest~(RTN), AWQ~\cite{lin2024awq}, GPTQ~\cite{frantar2022gptq}, and MBQ~\cite{li2025mbq} under W3A16, all employing channel-wise equalization and group-wise asymmetric quantization~(group size 128). For weight-activation quantization, we evaluate RTN, SmoothQuant~\cite{xiao2023smoothquant}, and MBQ under W4A8, also with channel-wise equalization. Following SmoothQuant, we use per-token symmetric quantization for activations and per-channel symmetric quantization for weights to utilize low-precision tensor cores.

\noindent \textbf{Datasets.}
To comprehensively assess the performance of our quantized models, we follow the LMMs-Eval~\cite{zhang2025lmms} protocol and evaluate on multiple vision--language benchmarks. In particular, MMMU~\cite{yue2024mmmu} and ScienceQA~\cite{lu2022learn} are used to test visual reasoning, VizWiz~\cite{gurari2018vizwiz} to examine real-world perception, and ChartQA~\cite{masry2022chartqa} and AI2D~\cite{kembhavi2016diagram} to evaluate the understanding of structured visual information.

\begin{table*}[htbp]
  \centering
  \renewcommand{\arraystretch}{1}
  \resizebox{0.94\textwidth}{!}{
    \begin{tabular}{ccccccccc}
      \toprule[1.3pt]
      \textbf{Model} & \textbf{Bitwidth} & \textbf{Method} & \textbf{VizWiz} & \textbf{MMMU} & \textbf{ChartQA} & \textbf{AI2D} & \textbf{ScienceQA} & \textbf{Avg.} \\
      \midrule
      \multirow{10}{*}{LLaVA-onevision-7B}
        & FP16  & -      & 60.41  & 49.22 & 80.04 & 81.31 & 95.88 & 73.37 \\
      \cmidrule(lr){2-9}
      ~ & \multirow{5}{*}{W3A16}
           & RTN        & \underline{59.12} & 43.67 & 68.88 & \underline{78.92} & 94.55 & 69.03 \\
      ~ & ~ & GPTQ      & 54.87 & 42.33 & 73.72 & 76.81 & 92.12 & 67.97 \\
      ~ & ~ & AWQ      & 58.65 & 42.89 & 74.08 & 77.92 & 82.20 & 67.15 \\
      ~ & ~ & MBQ       & 57.99 & \underline{44.00} & \underline{76.84} & 78.47 & \underline{94.89} & \underline{70.44} \\
      ~ & ~ & \textbf{QIG~(Ours)}       & \textbf{62.82} & \textbf{45.78} & \textbf{77.20} & \textbf{79.11} & \textbf{95.29} & \textbf{72.04}  \\
      \cmidrule(lr){2-9}
      ~ & \multirow{4}{*}{W4A8}
           & RTN       & 58.10 & 42.89 & 71.00 & 77.82 & 94.10 & 68.78 \\
      ~ & ~ & SQ        & 55.67 & 42.00 & 66.28 & 77.20 & 93.51 & 66.93 \\
      ~ & ~ & MBQ       & \underline{58.13} & \underline{44.78} & \textbf{74.92} & \underline{78.27} & \textbf{94.70} & \underline{70.16} \\
      ~ & ~ & \textbf{QIG~(Ours)}       & \textbf{59.10} & \textbf{45.00} & \underline{74.52} & \textbf{78.30} & \underline{94.25} & \textbf{70.23} \\
      
      \midrule

      \multirow{10}{*}{InternVL2-8B}
        & FP16  & -      & 60.86  & 48.56 & 82.64 & 82.42 & 97.07 & 74.31  \\
      \cmidrule(lr){2-9}
      ~ & \multirow{5}{*}{W3A16}
           & RTN       & 55.95 & 43.89 & 79.24 & \textbf{80.51} & \textbf{96.28} & 71.17 \\ 
      ~ & ~ & GPTQ      & \textbf{59.79} & 43.11 & 76.40 & 76.65 & 94.30 & 70.05  \\
      ~ & ~ & AWQ      & 58.14 & 45.56 & 74.42 & 79.47 & 95.88 & 70.70  \\
      ~ & ~ & MBQ       & 59.33 & \underline{46.02} & \textbf{80.04} & 79.66 & 95.93 & \underline{72.20} \\
      ~ & ~ & \textbf{QIG~(Ours)}       & \underline{59.55} & \textbf{46.22} & \textbf{80.04} & \underline{79.73} & \underline{96.03} & \textbf{72.31} \\
      \cmidrule(lr){2-9}
      ~ & \multirow{4}{*}{W4A8}
           & RTN       & 56.68 & 43.00 & \textbf{78.96} & 79.02 & 96.22 & 70.80 \\
      ~ & ~ & SQ        & 55.56 & 44.78 & 77.96 & 76.59 & 95.88 & 70.15 \\
      ~ & ~ & MBQ       & \underline{57.36} & \underline{45.67} & 78.00 & \underline{79.47} & \underline{96.38} & \underline{71.38} \\
      ~ & ~ & \textbf{QIG~(Ours)}       & \textbf{58.33} & \textbf{47.33} & \underline{78.16} & \textbf{79.63} & \textbf{96.73} & \textbf{72.04} \\
      \midrule

      \multirow{10}{*}{Qwen2-VL-7B}
        & FP16  & -      & 68.34 & 51.22 & 81.40 & 80.12 & 85.03 & 73.22 \\
      \cmidrule(lr){2-9}
      ~ & \multirow{5}{*}{W3A16}
           & RTN       & 65.02 & 44.67 & 73.64 & 76.33 & 81.06 & 68.14 \\
      ~ & ~ & GPTQ      & \textbf{67.73} & 44.44 & 76.20 & 74.87 & \underline{81.76} & 69.00 \\
      ~ & ~ & AWQ      & 66.24 & 45.89 & 77.08 & 77.53 & 81.01 & 69.56 \\
      ~ & ~ & MBQ       & 66.62 & \underline{46.48} & \textbf{79.18} & \underline{77.81} & \textbf{81.85} & \underline{70.15} \\
      ~ & ~ & \textbf{QIG~(Ours)}       & \underline{67.12} & \textbf{47.11} & \underline{77.76} & \textbf{77.88} & 81.61 & \textbf{70.30} \\
      
      \cmidrule(lr){2-9}
      
      ~ & \multirow{4}{*}{W4A8}
           & RTN       & 58.71 & \underline{45.44} & 74.16 & \underline{77.01} & \underline{79.62} & 66.99 \\
      ~ & ~ & SQ        & 47.60 & 43.78 & 70.88 & 76.07 & 78.98 & 63.46 \\
      ~ & ~ & MBQ       & \textbf{60.17} & 44.89 & \textbf{76.92} & 76.49 & 78.93 & \underline{67.48} \\
      ~ & ~ & \textbf{QIG~(Ours)}       & \underline{58.85} & \textbf{46.00} & \underline{76.68} & \textbf{77.17} & \textbf{80.17} & \textbf{67.77} \\
      \bottomrule[1.3pt]
    \end{tabular}
  }
  \caption{Overall comparison of full-precision and post-training quantization methods on three representative LVLMs under W3A16 and W4A8. RTN and SQ are naive PTQ baselines, MBQ is the modality-balanced baseline, and \textbf{QIG} is the proposed fine-grained quantization method. \textbf{Bold} numbers indicate the best performance, and \underline{underlined} numbers indicate the second best in each column.}
  \label{tab:main-result}
  \vspace{-8pt}
\end{table*}

\subsection{Main Results}
Tab.~\ref{tab:main-result} reports the performance of different PTQ methods on three representative LVLMs under both weight-only~(W3A16) and weight–activation~(W4A8) quantization. 

\noindent \textbf{Generic LLM PTQ methods underperform naive RTN on LVLMs.}
Across all three models, the naive RTN baseline already causes a moderate drop~(about 4\% on average) compared with FP16, indicating that 3-bit weight quantization is non-trivial for LVLMs. However, GPTQ and SmoothQuant~(SQ), which are strong PTQ methods for pure LLMs, do not reliably improve performance in this multimodal setting. Under W3A16, GPTQ often lags behind RTN in terms of average accuracy~(\eg., LLaVA-onevision-7B and InternVL2-8B), and under W4A8, SQ is consistently worse than RTN on all three models. In other words, directly applying PTQ methods designed for LLMs to LVLMs, while ignoring cross-modal statistical characteristics, may perform no better than simple round-to-nearest and can even degrade performance. This observation underscores the importance of leveraging multimodal information when designing quantization strategies for LVLMs.

\noindent \textbf{Fine-grained token-level sensitivity weighting beyond modality-level quantization.}
Modality-aware quantization provides a strong starting point for quantizing LVLMs. The MBQ baseline reweights the reconstruction errors of the \emph{vision} and \emph{language} modalities to alleviate their inherent imbalance during quantization. As a result, MBQ achieves consistent improvements of about 1\% on average over RTN and GPTQ across three models and both bitwidths. However, modality-level balancing remains coarse, since tokens within the same modality can exhibit different sensitivities to quantization. This limitation motivates the fine-grained token-level sensitivity weighting proposed in our method.

To further address the limitations of modality-level sensitivity modeling, our method introduces fine-grained token-level sensitivity weighting. Across six quantized configurations, including three foundation models and two bitwidth settings, our method consistently achieves the highest average accuracy. Compared with MBQ, it brings an additional average gain of about 0.5\%. For example, on LLaVA-onevision-7B, the average accuracy improves from 70.44\% to 72.04\% under W3A16 and from 70.16\% to 70.23\% under W4A8. Similar steady improvements are observed on InternVL2-8B and Qwen2-VL-7B under both bitwidths. Moreover, across all benchmarks and quantized configurations in Tab~\ref{tab:main-result}, our method either achieves the best performance or remains the second best among all PTQ baselines. The gains are particularly clear on challenging benchmarks. On VizWiz and MMMU, our method surpasses MBQ by around 1\% on average, which suggests that token-level weighting better preserves sensitive visual and reasoning tokens. This improvement may stem from estimating token-wise sensitivity rather than using a single weight per modality, enabling finer control over token importance. Qualitative visualizations in the Appendix~\ref{sec:visualizations} show that, under the same quantization settings, our method yields more accurate answers than MBQ.

\begin{table}[t]
  \centering
  \resizebox{\linewidth}{!}{
    \begin{tabular}{cccccc}
    \toprule[1.3pt]
      \textbf{Model} & \textbf{Bitwidth} & \textbf{Method} & \textbf{ChartQA} & \textbf{MMMU} & \textbf{VizWiz}\\
      \midrule
      & FP16   & -    & 86.44 & 52.78 & 65.65 \\
      \cmidrule(lr){2-6}
      & \multirow{2}{*}{W4A8}  & MBQ  & 84.44 & 49.78 & 63.51 \\
      InternVL2-26B&& \textbf{Ours} & 85.24 & 50.22 &  63.91\\
      \cmidrule(lr){2-6}
      & \multirow{2}{*}{W3A16} & MBQ  & 84.48 & 51.67 & 63.33 \\
      &                        & \textbf{Ours} & 85.12 & 50.89 & 64.14\\
      \bottomrule[1.3pt]
    \end{tabular}
  }
  \caption{Quantization on InternVL2-26B: MBQ vs. Ours under W3A16/W4A8.}
  \label{tab:larger_model}
  \vspace{-10pt}
\end{table}

\noindent \textbf{Scaling to Larger Models.}
To assess whether the proposed fine-grained post-training quantization scales to larger LVLMs, we further apply it to InternVL2-26B and compare it with MBQ under both W4A8 and W3A16 configurations. As shown in Tab.~\ref{tab:larger_model}, our method yields clear gains over MBQ on ChartQA and VizWiz for both bitwidth settings, while maintaining comparable performance on MMMU. Under the W4A8 configuration, our approach recovers most of the FP16 accuracy, keeping the performance drop within 3\% on all benchmarks. Even under the more aggressive W3A16 setting, our method still surpasses MBQ on ChartQA and VizWiz and remains within 2\% of the FP16 model across all tasks, despite using 3-bit weights. These results demonstrate that the proposed fine-grained quantization strategy scales reliably to LVLMs with tens of billions of parameters and can be deployed at larger model sizes without incurring substantial performance degradation.

\subsection{Ablation Study and Further Analysis}
We conduct ablation studies to examine the effectiveness of fine-grained quantization, framework generality, and quantization efficiency. The results show that each design component contributes measurable performance gains while introducing negligible additional computational overhead. Additional experimental results, including more ablation studies, are presented in the Appendix~\ref{sec:more_experimental}.

\begin{table}[t]
  \centering
  \small
  \begin{tabular}{cccc}
    \toprule[1.3pt]
    \textbf{baseline} & \textbf{Attribution objective} & \textbf{ChartQA} & \textbf{VizWiz}\\
    \midrule
    $0$           & $f(x)$ 
                  & 73.87 & 61.73 \\
    $0$           & $f(x) - f(0)$ 
                  & 74.30 & 62.31 \\
    $x^q$ & $f(x)$ 
                  & 74.12 & 61.52 \\
    $x^q$ & $f(x) - f(x^q)$
                  & \textbf{74.52} & \textbf{62.82} \\
    \bottomrule[1.3pt]
  \end{tabular}
  \caption{
    Ablation of the integrated-gradients configuration for token-wise
    sensitivities, varying the reference baseline $x'$~(0 vs.\ $x^q$)
    and attribution objective~(task output $f(x)$ vs.\ quantization-error
    outputs $f(x) - f(0)$ or $f(x) - f(x^q)$).
    Results are on LLaVA-onevision-7B with W4A8; the last row
    is our QIG formulation and performs best~(higher is better).
  }
  \label{tab:baseline_objective_ablation}
  \vspace{-8pt}
\end{table}

\noindent \textbf{Sensitivity Ablation of Fine-Grained Quantization.}
\label{sec:abla_fg_effect}
In Sec.~\ref{sec:main_method}, we present our quantization-aware integrated gradients, which depart from the standard formulation in two key aspects: the choice of reference baseline and the scalar objective whose gradients are integrated along the path. To evaluate the contribution of these components to fine-grained quantization, we perform an ablation study over both the baseline and the objective used to compute token-wise sensitivities. We evaluate four configurations of Integrated Gradients on LLaVA-onevision-7B under the W4A8 setting, and report downstream accuracies on ChartQA and VizWiz in Tab.~\ref{tab:baseline_objective_ablation}. We ablate over two choices: the baseline $x' \in \{0, x^q\}$ and the attribution objective $g(x) \in \{f(x), f(x) - f(0), f(x) - f(x^q)\}$. Our QIG formulation corresponds to the configuration with baseline $x' = x^q$ and objective $g(x) = f(x) - f(x^q)$.

From the results in Tab.~\ref{tab:baseline_objective_ablation}, we observe that both components of QIG contribute to the final performance. Under the zero baseline, switching the objective from the task output $f(x)$ to the error $f(x) - f(0)$ already yields consistent gains on ChartQA~(+0.43\%) and VizWiz~(+0.58\%). Changing the baseline from $0$ to $x^q$ while keeping the task-output objective provides a small improvement on ChartQA~(73.87\% to 74.12\%) but slightly hurts VizWiz~(61.73\% to 61.52\%). In contrast, combining the quantized baseline with the error objective $f(x) - f(x^q)$ leads to the best results on both datasets, achieving 74.52\% on ChartQA and 62.82\% on VizWiz. These trends indicate that integrating the quantized input and explicitly attributing the quantization error are both important for obtaining reliable token-wise sensitivities for post-training quantization.

\noindent \textbf{Combine Fine-Grained Quantization with GPTQ. }
\label{sec:abla_igptq_quant}
To further demonstrate the generality of our fine-grained quantization strategy, we incorporate it into the GPTQ framework~\cite{frantar2022gptq}, which minimizes layer-wise reconstruction error through second-order approximation using the Hessian matrix $H = X^{\top} X$. In our adaptation, we introduce a token-aware modification by replacing the Hessian with $H' = X^{\top} \Lambda X$, where $\Lambda = diag(\lambda_1, \lambda_2, \ldots, \lambda_T)$ represents the token importance coefficients derived from our fine-grained attribution mechanism. This reweighting allows GPTQ to emphasize activations from quantization-sensitive tokens while maintaining the overall optimization structure. Notably, the modification requires no additional calibration data and incurs negligible computation overhead, making it a plug-and-play enhancement to standard GPTQ.

\begin{table}[t]
  \centering
  \resizebox{\linewidth}{!}{
      \begin{tabular}{cccccc}
        \toprule[1.5pt]
        \textbf{Model} & \textbf{Bitwidth} & \textbf{Method} & \textbf{ChartQA} & \textbf{AI2D} & \textbf{VizWiz} \\
        \midrule
        \multirow{3}{*}{\shortstack{\\[3pt] LLaVA\\-onevision\\-7B}}
        & FP16 & - & 80.04 & 81.31 & 60.41 \\
        \cmidrule(lr){2-6}
        & \multirow{2}{*}{W3A16} & GPTQ & 73.72& 76.81 & 54.87 \\
        &  & \textbf{+ Ours} &74.12 & 76.65 & 56.95\\
    
        \midrule
        \multirow{3}{*}{\shortstack{\\[6pt] InternVL2\\-8B}}
        & FP16 & - & 82.64 & 82.42 & 60.86\\
        \cmidrule(lr){2-6}
        & \multirow{2}{*}{W3A16} & GPTQ & 76.40 & 76.65 & 59.79 \\
        &  & \textbf{+ Ours} & 78.12  & 78.47 & 60.57 \\
        \bottomrule[1.5pt]
      \end{tabular}}
  \caption{Results of combining our fine-grained quantization with GPTQ under the W3A16.}
  \label{tab:gptq}
  \vspace{-10pt}
\end{table}

As shown in Tab.~\ref{tab:gptq}, combining fine-grained weighting with GPTQ consistently improves quantization performance on both LLaVA-onevision-7B and InternVL2-8B under the W3A16 setting. For instance, our fine-grained variant achieves 56.95\% on VizWiz for LLaVA-onevision-7B, surpassing vanilla GPTQ by 2.08\%, and brings notable gains on ChartQA and AI2D for both models. This strongly demonstrates the effectiveness and scalability of our method, highlighting the advantages and necessity of fine-grained quantization.

\noindent \textbf{Quantization Efficiency.}
\label{sec:quantization-effciency}
To evaluate the practical efficiency of our fine-grained quantization, we measure the total quantization time required to process each model under different configurations. For comparison, we include the baseline MBQ~\cite{li2025mbq} and the perturbation-based Leave-One-Out strategy. The metric reports the total wall-clock GPU hours spent during the calibration and scale-search stages, including activation collection and layer-wise optimization. 

\begin{table}[t]
  \centering
  \resizebox{\linewidth}{!}{
  \begin{tabular}{lccc}
    \toprule[1.3pt]
    \multirow{2}{*}{Model Size} & \multicolumn{3}{c}{GPU Hours} \\
    \cmidrule(lr){2-4}
     & MBQ & Leave One Out & \textbf{Ours} \\
    \midrule
    InternVL2-8B  & 0.55 &2.07~(+91 min) &  0.58~(+2.0 min) \\
    InternVL2-26B & 0.95 & 4.20~(+ 195 min)&  0.99~(+2.5 min) \\
    \bottomrule[1.3pt]
  \end{tabular}}
  \caption{Quantization time~(in GPU hours) of differnet models using a single A800 80GB GPU. Fine-Grained Quantization incurs negligible overhead compared to baseline methods.}
  \label{tab:efficiency}
\end{table}

As shown in Tab.~\ref{tab:efficiency}, our fine-grained method introduces only negligible overhead compared to MBQ, approximately two additional minutes for both InternVL2-8B and InternVL2-26B, while achieving consistent accuracy improvements. In contrast, the Leave-One-Out approach, while also effective in measuring quantization error at the token level, incurs high computational cost, consuming about 3--4 $\times$ more GPU time due to repeated forward passes for each token perturbation. These results verify that the proposed fine-grained quantization effectively balances interpretability, accuracy, and computational efficiency, making it effective across different architectures and scalable to larger LVLMs in real deployment scenarios.

\section{Conclusion}
In this work, we revisited post-training quantization for LVLMs and showed that conventional modality-level sensitivity modeling is fundamentally insufficient. Our analysis of cross-token interactions reveals that tokens within the same modality exhibit substantial differences in quantization sensitivity. To bridge this granularity gap, we introduced Quantization-aware Integrated Gradients~(QIG), an attribution-based framework that decomposes the quantization error between full-precision and quantized models into token-level contributions. By integrating from the quantized input and applying robust clipping, QIG provides stable importance scores that effectively guide fine-grained quantization. Our approach outperforms existing PTQ methods across diverse benchmarks. Under 3-bit weight-only quantization, it improves the average accuracy of LLaVA-onevision-7B by 1.60\%, reducing the gap to its full-precision counterpart to just 1.33\%. We believe this token-aware, attribution-guided view of quantization offers a practical path toward deploying compact yet reliable LVLMs and motivates future work on unified, token-level compression in real-world systems.

\clearpage 

{
    \small
    \bibliographystyle{ieeenat_fullname}
    \bibliography{main}
}

\clearpage
\setcounter{page}{1}
\setcounter{figure}{0}
\setcounter{table}{0}
\renewcommand{\thefigure}{A\arabic{figure}}
\renewcommand{\thetable}{A\arabic{table}}
\maketitlesupplementary
\appendix

\section{Proof of Quantization-Aware Integrated Gradients Completeness}
\label{sec:qig-completeness}
We denote the input as $x = [x_1, \ldots, x_T]$, where each token embedding
$x_i \in \mathbb{R}^d$. Thus the full input lies in $\mathbb{R}^{T \times d}$. 
Using the definition of QIG in Eq.~\ref{eq:qig}, the attribution for
the $i$-th token is defined as:
\begin{equation}
\small
\mathrm{QIG}_i(x)
  = (x_i - x_i^{q})
  \int_0^1 
  \frac{\partial \left( f(x_{\alpha}, w) - f(x_{\alpha}, w^{q}) \right)}
       {\partial x_i}\, d\alpha,
\end{equation}
where $x_{\alpha} = x^{q} + \alpha (x - x^{q})$ is the linear interpolation
between the quantized input $x^{q}$ and the original input $x$.

To simplify the notation, we define the quantization-error function as:
\begin{equation}
G(x) = f(x, w) - f(x, w^{q}).
\end{equation}
Under this definition, QIG becomes standard Integrated Gradients~(IG) applied
to $G(\cdot)$ with baseline $x^{q}$:
\begin{equation}
\mathrm{QIG}_i(x)
  = (x_i - x_i^{q})
  \int_0^1 
  \frac{\partial G(x_{\alpha})}{\partial x_i}\, d\alpha.
\end{equation}

\paragraph{Completeness.}
Consider the interpolation path $\gamma(\alpha) = x_{\alpha}=x^{q} + \alpha (x - x^{q})$.
Since the path is linear, the derivative with respect to $\alpha$ can be written as:
\begin{equation}
\frac{\partial x_{\alpha}}{\partial\alpha}
   = x - x^{q}.
\end{equation}
Applying the chain rule to $G(\gamma(\alpha))$ yields:
\begin{equation}
\begin{aligned}
\frac{\partial}{\partial\alpha} G(x_{\alpha})
 &= \nabla_x G(x_{\alpha})^{\top} 
    \frac{\partial x_{\alpha}}{\partial\alpha} \\[4pt]
 &= \nabla_x G(x_{\alpha})^{\top}(x - x^{q}) \\[4pt]
 &= \sum_{i=1}^T (x_i - x_i^{q})
     \frac{\partial G(x_{\alpha})}{\partial x_i},
\end{aligned}
\end{equation}
which shows that the weighted coordinate-wise gradients in QIG correspond to
the directional derivative of $G$ along the interpolation path.

Integrating both sides from $\alpha=0$ to $1$, and using
$G(\gamma(0)) = G(x^{q})$ and $G(\gamma(1)) = G(x)$, the fundamental theorem
of calculus gives:
\begin{equation}
\begin{aligned}
G(x) - G(x^{q})
 &= \int_0^1 
    \frac{\partial}{\partial\alpha} G(x_{\alpha})\, d\alpha \\[4pt]
 &= \sum_{i=1}^T 
    (x_i - x_i^{q})
    \int_0^1 
    \frac{\partial G(x_{\alpha})}{\partial x_i}\, d\alpha.
\end{aligned}
\end{equation}
Recognizing the definition of $\mathrm{QIG}_i(x)$, we obtain the completeness
property:
\begin{equation}
\begin{aligned}
\sum_{i=1}^T \mathrm{QIG}_i(x)
 &= G(x) - G(x^{q}) \\[2pt]
 &= \big[f(x, w) - f(x, w^{q})\big] \\
 &\quad - \big[f(x^{q}, w) - f(x^{q}, w^{q})\big].
\end{aligned}
\end{equation}

\paragraph{Discussion.}
When the baseline satisfies $G(x^{q}) = 0$~(\eg, when $f(x^{q}, w) = f(x^{q}, w^{q})$), the completeness relation simplifies to:
\begin{equation}
\sum_{i=1}^T \mathrm{QIG}_i(x)
 = f(x, w) - f(x, w^{q}),
\end{equation}
which mirrors the classical IG completeness property. In practice, post-processing of $\mathrm{QIG}$ values~(\eg, clipping or interquartile-range filtering) may slightly break strict algebraic completeness while improving numerical stability and visualization quality.

\section{More Implementation Details}
\label{app:imple-detail}
\noindent \textbf{QIG objective implementation.}
Let $x \in \mathbb{R}^{B \times T \times H}$ be the pre-residual activation of the current block, and let $x^{q}$ be its quantized version. The block outputs are denoted as $y_{\mathrm{fp}} = f(x,w)$ and $y_{\mathrm{q}} = f(x,w^{q})$. We define the per-token quantization distortion error as:
\[
E_{b,t}(x)
  = \frac{1}{H}\sum_{h=1}^{H}
    \bigl|\,(y_{\mathrm{fp}} - y_{\mathrm{q}})_{b,t,h}\,\bigr|,
\qquad
E(x)\in\mathbb{R}^{B\times T}.
\]

To obtain QIG attributions, we approximate the gradients of this quantization distortion loss $E_{b,t}(x)$ using \textbf{32-step integrated gradients}~\cite{sundararajan2017axiomatic}. Specifically, we integrate along the straight-line path from the baseline $x^{q}$ to the input $x$, defined as $x(\alpha)=x^{q}+\alpha(x-x^{q})$ for $\alpha \in [0, 1]$. Crucially, this computation is performed directly on the difference function $\|f(x) - f_q(x)\|$, \textbf{without separately computing or subtracting gradients} from the full-precision and quantized models individually. This follows the construction described in Appendix~\ref{sec:qig-completeness}, where $E_{b,t}(x)$ serves as the scalar target function for attribution.

\paragraph{Quantization Formats.}
We adopt uniform integer quantization for all experiments, and quantize both the weights $W$ and the input activations $X$ of each linear layer. For a given tensor $T$ and bit-width $b$, we denote its quantized integer representation by $Q(T) \in \mathbb{Z}^{b}$ and the corresponding dequantized value by $\hat{T}$.

For weight-only quantization, we apply asymmetric, group-wise quantization to the weight matrix $W$. Each row of $W$ is partitioned into non-overlapping groups of size $128$, and for each group $g$ we compute a scale $s_g$ and zero-point $z_g$ from the group-wise minimum and maximum:

\begin{align}
s_g &= \frac{\max(W_g) - \min(W_g)}{2^b - 1}, \\
z_g &= \operatorname{round}\left(-\frac{\min(W_g)}{s_g}\right).
\end{align}

The integer weights are then obtained as:
\begin{align}
Q(W_g) = \operatorname{clip}\Bigl(\operatorname{round}\!\bigl(W_g / s_g\bigr) + z_g,\; 0,\; 2^b - 1\Bigr),
\end{align}
and the dequantized weights are $\hat{W}_g = s_g \bigl(Q(W_g) - z_g\bigr)$. We primarily use $b \in \{3,4\}$, which we denote as $W3$ and $W4$.

For weight–activation quantization, we use symmetric quantization for both weights and activations. Given a tensor $T$ and bit-width $b$, we define:
\begin{equation}
\small
\begin{gathered}
s_T = \frac{\max(|T|)}{2^{b-1}-1},\\
Q(T) = \operatorname{clip}\bigl(\operatorname{round}(T / s_T),
        -2^{b-1}, 2^{b-1}-1\bigr),
\end{gathered}
\end{equation}
and $\hat{T} = s_T \, Q(T)$. In this setting we write $W x A y$ to indicate $x$-bit weight and $y$-bit activation quantization, \eg, $W4A8$ for $4$-bit weights and $8$-bit activations. Unless otherwise stated, the group size for weight quantization is fixed to $128$.

\section{More Experimental Results}
\label{sec:more_experimental}
\noindent \textbf{Effectiveness of IQR-Based Clipping.}
To more comprehensively evaluate the robustness benefits introduced by our IQR-based clipping strategy, we conduct an ablation study comparing four sensitivity stabilization variants: (1) No Clipping, which directly uses raw token-level sensitivities; (2) Top-5 Zero, which suppresses the five largest sensitivity values by setting them to zero; (3) Top-5 Average, which replaces the five largest sensitivities with the global mean computed over all token sensitivities; and (4) our full IQR Clipping method, which attenuates extreme values using statistically grounded interquartile-range thresholds.

As shown in Tab.~\ref{tab:iqr_ablation}, all clipping strategies improve performance relative to the raw-sensitivity baseline, highlighting the importance of controlling outlier sensitivities. Notably, modifying the importance allocation of only five tokens already leads to clear performance differences, underscoring the necessity of fine-grained, token-level importance estimation. Among all variants, our IQR-based approach achieves the best results across VizWiz, MMMU, and ScienceQA, demonstrating that the observed gains originate not merely from simple top-value replacement, but from a distribution-aware clipping mechanism that more effectively stabilizes the sensitivity distribution.

\begin{table}[t]
  \renewcommand{\arraystretch}{1.3}
  \centering
  \small
  \begin{tabular}{lcccc}
    \toprule[1.3pt]
    Method & VizWiz & MMMU  & ScienceQA\\
    \midrule
    No Clipping 
        & 54.32 & 41.37 & 93.28 \\
    Top5 zero
        & 57.20& 43.56 & 94.10\\                
    Top5 average
        & 57.25& 44.78 & 94.18\\
    IQR Clipping~(Ours)
        & \textbf{59.10} & \textbf{45.00} & \textbf{94.25} \\
    \bottomrule[1.3pt]
  \end{tabular}
  \caption{
    Ablation on sensitivity stabilization strategies.
    Our IQR Clipping achieves the best overall performance
    on LLaVA-OneVision-7B under W4A8 quantization.
  }
  \label{tab:iqr_ablation}
\end{table}

\noindent \textbf{Extension to Large Language Models~(LLMs).}
To verify that our method’s effectiveness stems from accurately measuring token-level sensitivity rather than serving as a simple modality-related replacement, we further extend our approach to LLMs. Tab.~\ref{tab:llm} reports results with quantized LLaMA-2 on several standard language understanding benchmarks, including perplexity~(PPL), PIQA for physical commonsense reasoning, ARC-e/ARC-c for scientific question answering, and MMLU for multi-domain knowledge understanding. As shown in Tab.~\ref{tab:llm}, our fine-grained quantization method not only performs strongly on LVLMs but also achieves notable improvements when applied to LLMs. Specifically, by leveraging QIG to model token-level sensitivity, we attain superior quantization performance across different modalities and model types. This capability to capture fine-grained token sensitivity makes our method highly versatile, enabling consistent performance gains across various large-scale pre-trained models, including both multimodal and unimodal settings.

\begin{table}[t]
\centering
\resizebox{\linewidth}{!}{
\begin{tabular}{lccccc}
\toprule[1.3pt]
 & PPL$\downarrow$ & PIQA $\uparrow$& ARC-e$\uparrow$ & ARC-c $\uparrow$ & MMLU $\uparrow$\\
\midrule
GPTQ      & 6.24 & 75.46 & 67.00 & \textbf{40.10} & 30.05 \\
\textbf{+ Ours}    & \textbf{6.19} & \textbf{75.95} & \textbf{67.17} & 39.85 & \textbf{32.01} \\
\bottomrule[1.3pt]
\end{tabular}}
\caption{Comparison of GPTQ and Our Fine-Grained Quantization on LLaMA-2-7B~(3bit).}
\vspace{-8pt}
\label{tab:llm}
\end{table}

\noindent \textbf{Robustness with OCR-Specific Calibration.}
To address concerns regarding the method's adaptability to domain-specific challenges, we evaluate our approach using an OCR-focused calibration set derived from InfoVQA data.  Tab.~\ref{tab:qwen2vl_ocr_quant} reports the performance on Qwen2-VL-7B under W4A8 quantization across three OCR-intensive benchmarks: DocVQA, ChartQA, and OCRBench. As shown in the table, our method consistently outperforms the MBQ baseline across all calibration sizes~(128 and 256 samples). Specifically, with only 128 calibration samples, our approach achieves an average improvement of \textbf{3.52\%} over MBQ, with notable gains of +4.12\% on DocVQA and +6.20\% on OCRBench. Even as the calibration size increases to 256, our method maintains a significant lead~(avg. +3.38\%). These results demonstrate that our token-level sensitivity modeling effectively captures critical features for text-rich visual understanding, ensuring robustness even when calibration data is limited or domain-specific.

\begin{table}[t]
  \centering
  \small
  \resizebox{\linewidth}{!}{
    \begin{tabular}{ccccccc}
    \toprule[1.3pt]
    Bitwidth & Calib. Size & Method & DocVQA & ChartQA & OCRBench & Avg. \\
    \midrule
    \multirow{4}{*}{W4A8}
      & \multirow{2}{*}{128} 
      & MBQ   & 84.48 & 77.28 & 70.60 & 77.45 \\
      &       & \textbf{Ours} & \textbf{88.60} & \textbf{77.52} & \textbf{76.80} & \textbf{80.97} \\
    \cmidrule(lr){2-7}
      & \multirow{2}{*}{256}
      & MBQ   & 84.87 & 76.68 & 71.50 & 77.68 \\
      &       & \textbf{Ours} & \textbf{89.13} & \textbf{77.04} & \textbf{77.00} & \textbf{81.06} \\
    \bottomrule[1.3pt]
    \end{tabular}
  }
  \caption{\footnotesize{Results on Qwen2-VL-7B using OCR-specific calibration data. Our method shows significant robustness improvements over MBQ in text-rich scenarios.}}
  \label{tab:qwen2vl_ocr_quant}
\end{table}

\section{Visualizations}
\label{sec:visualizations}
\begin{figure*}[t]
    \centering
    \includegraphics[width=0.9\textwidth]{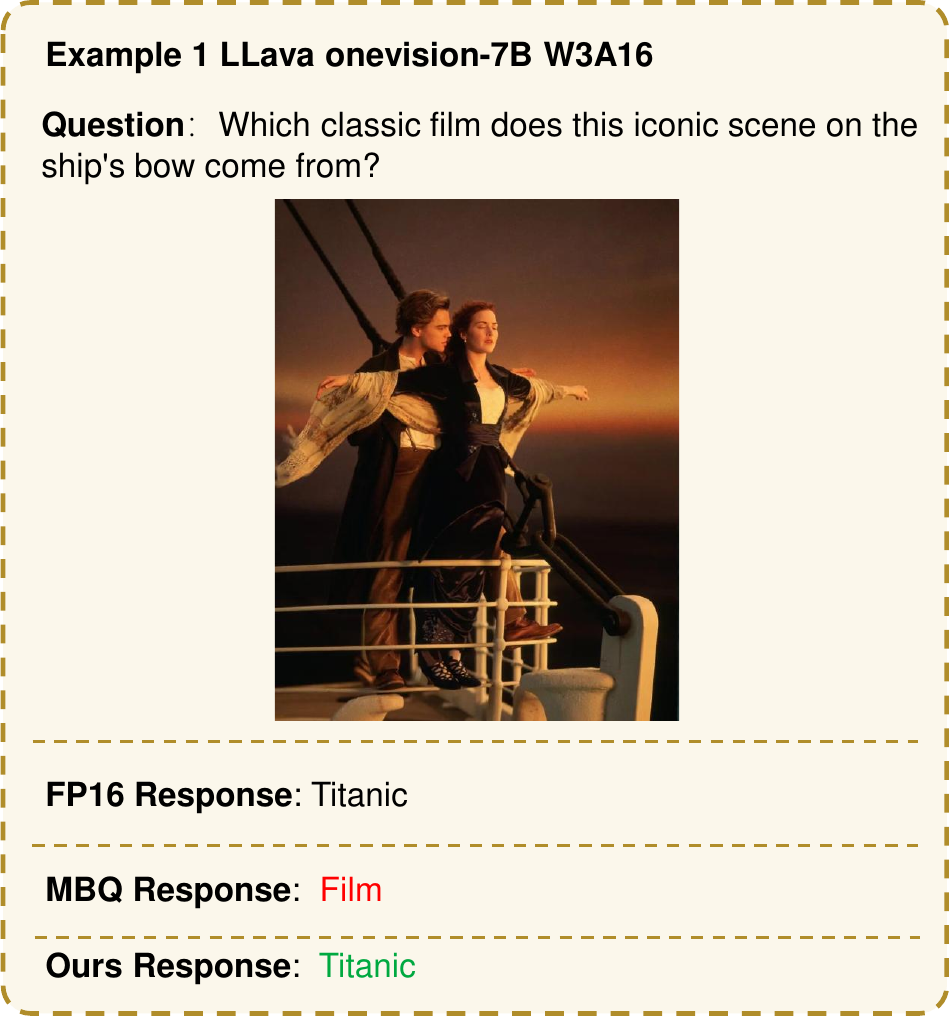}  
    \caption{The baseline fails to identify the film and produces an incomplete answer, whereas our fine-grained quantization successfully preserves the correct semantic prediction and matches the full-precision model.}

      \label{fig:v1}
    \vspace{-8pt}
\end{figure*}

\begin{figure*}[t]
    \centering
    \includegraphics[width=0.9\textwidth]{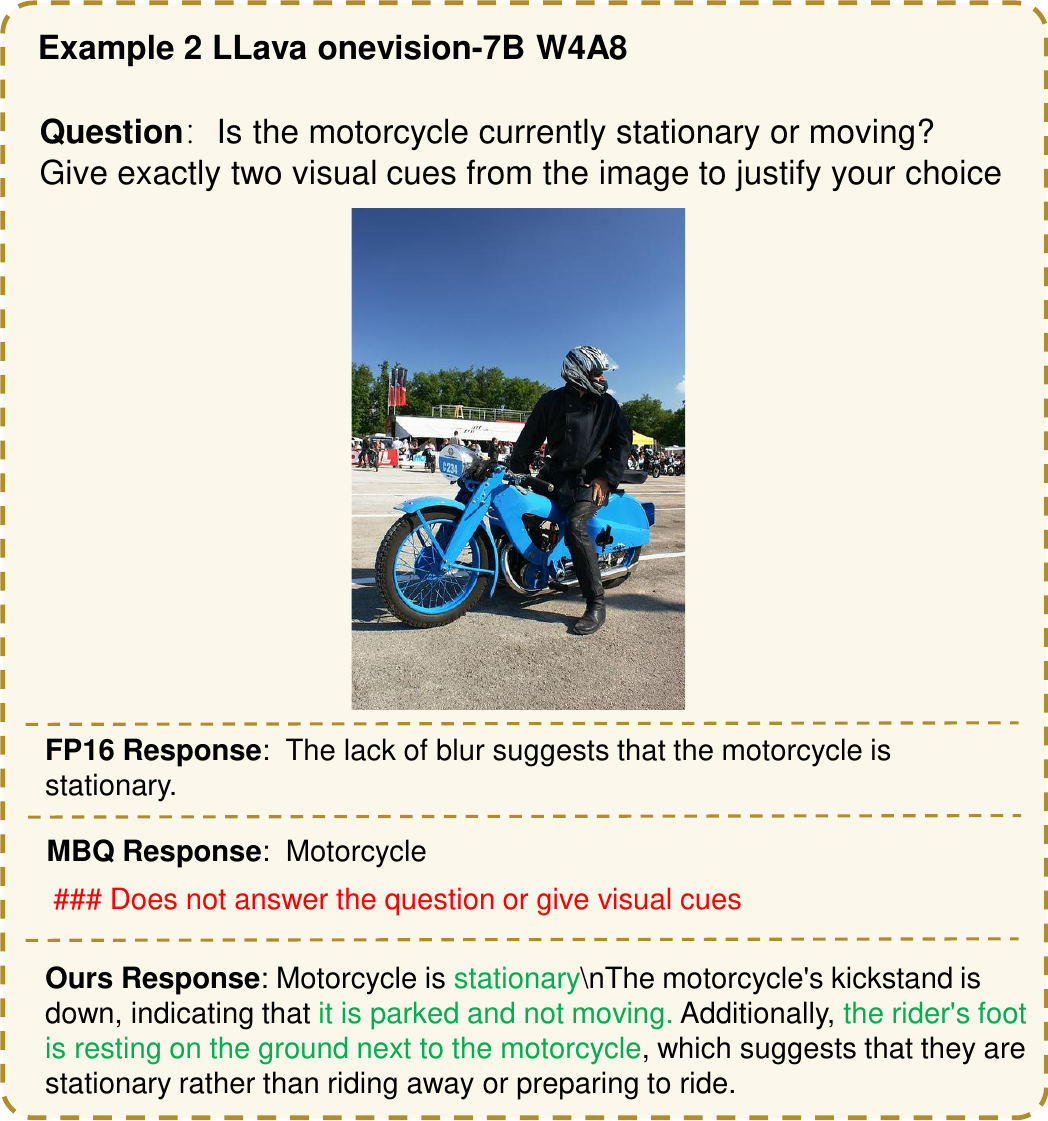}  
    \caption{The baseline fails to answer the question and provides no reasoning, whereas our fine-grained quantization preserves both correctness and detailed visual justification, closely matching the full-precision model.}

      \label{fig:v2}
    \vspace{-8pt}
\end{figure*}

\begin{figure*}[t]
    \centering
    \includegraphics[width=0.9\textwidth]{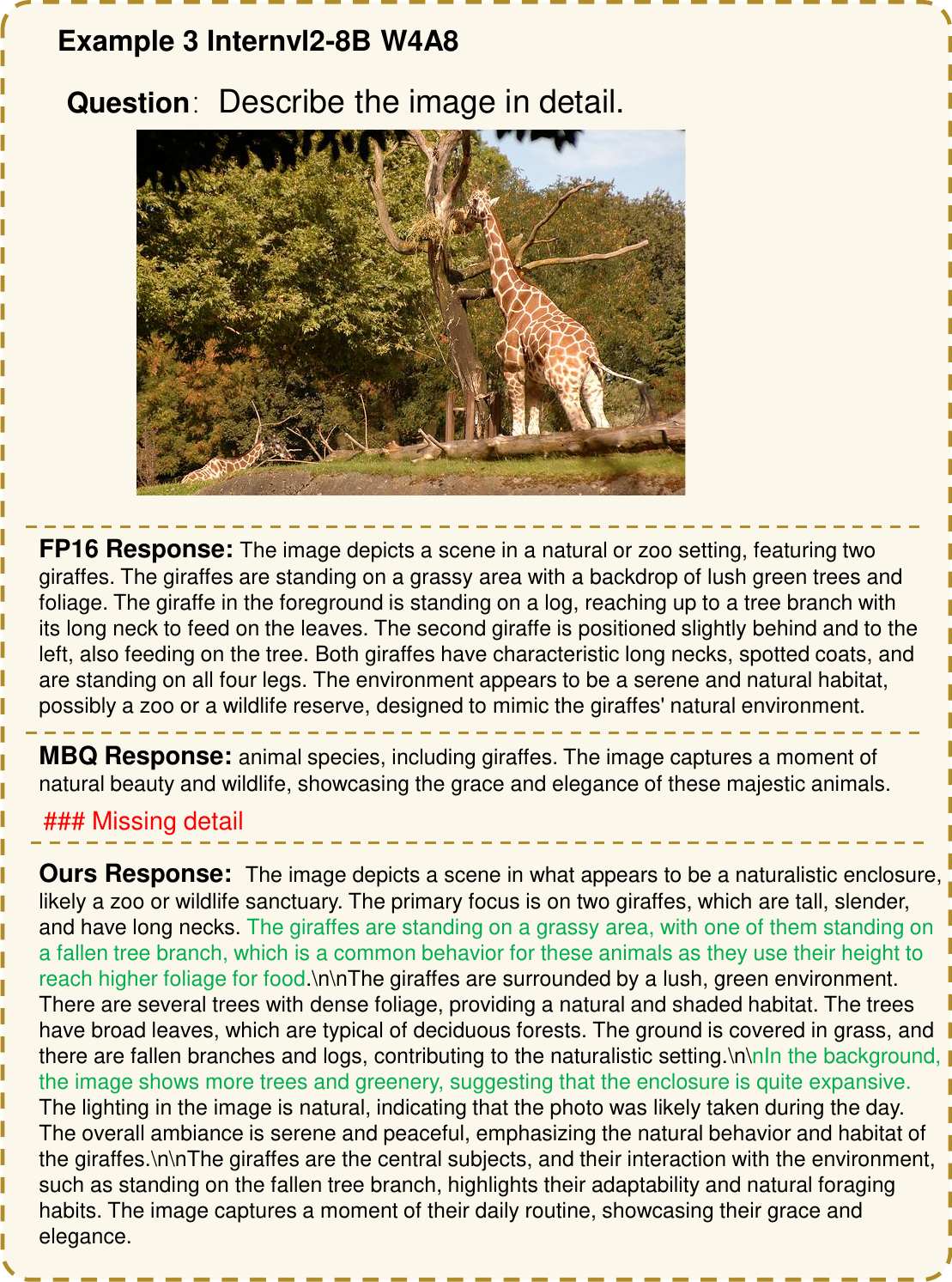}  
    \caption{The baseline provides only a minimal and overly generic description, missing most visual details, whereas our fine-grained quantization preserves rich scene understanding and produces a comprehensive description close to the full-precision model.}

      \label{fig:v3}
    \vspace{-8pt}
\end{figure*}

\begin{figure*}[t]
    \centering
    \includegraphics[width=0.9\textwidth]{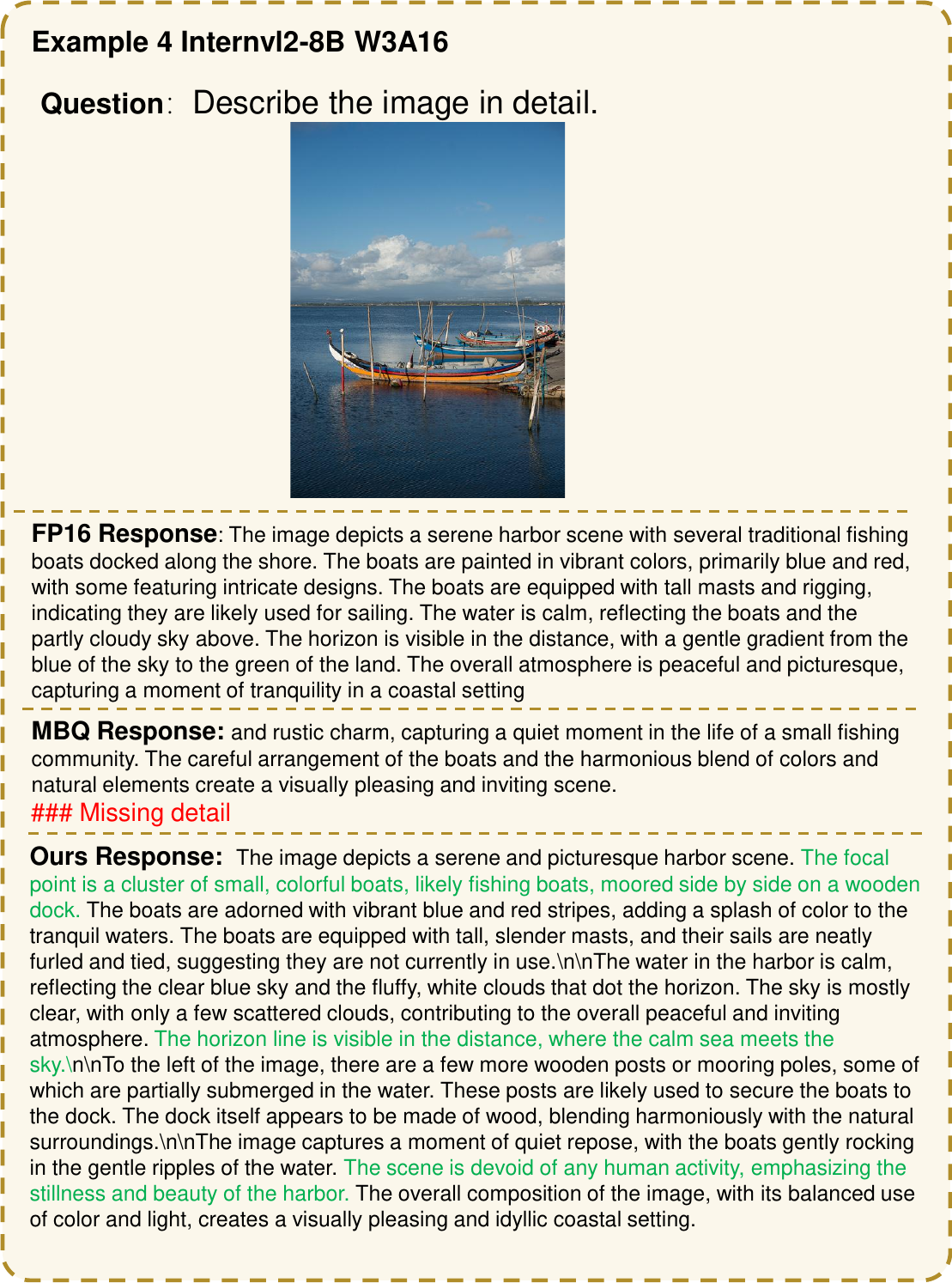}  
    \caption{The baseline produces an incomplete and overly generic description that misses key scene elements, whereas our fine-grained quantization preserves detailed coastal features and provides a rich interpretation closely aligned with the full-precision model.}
      \label{fig:v4}
    \vspace{-8pt}
\end{figure*}

In this section, we provide extended visualizations to further analyze the conversational outputs of vision–language models under different quantization schemes. The comparative results, visually  shown in Figs.~\ref{fig:v1}--\ref{fig:v4}, indicate that our proposed fine-grained quantization strategy enables the quantized model’s responses to better align with the calibration data, effectively reducing degradation in reasoning quality, visual detail retention, and linguistic coherence, thereby more clearly demonstrating its advantages over modality-based baseline methods.

\clearpage

\end{document}